\documentclass[lettersize,journal]{IEEEtran}
\usepackage{amsmath,amsfonts}
\usepackage{algorithmic}
\usepackage{algorithm}
\usepackage{array}
\usepackage[caption=false,font=normalsize,labelfont=sf,textfont=sf]{subfig}
\usepackage{textcomp}
\usepackage{stfloats}
\usepackage{url}
\usepackage{verbatim}
\usepackage{graphicx}
\usepackage{color}
\usepackage{cite}

\IEEEoverridecommandlockouts

\title{\LARGE \bf
Multi-Sensor Fusion-Based Mobile Manipulator Remote Control for Intelligent Smart Home Assistance}

\author{
Xiao Jin$^{1}$, Bo Xiao$^{2}$, Huijiang Wang$^{3}, $Wendong Wang$^{4}$,  Zhenhua Yu*$^{1}$$^{5}$\\
\thanks{*Corresponding author: Zhenhua Yu}%
\thanks{This work was partly supported by Capital Equipment funding of University of Aberdeen.}%
\thanks{$^{1}$ Dyson School of Design Engineering, Imperial College London, SW7 2DB, UK.
       }%
\thanks{$^{2}$ School of Robotics, Xi'an Jiaotong-Liverpool University (XJTLU), Suzhou, China, 215000.
        }

\thanks{$^{3}$Department of Engineering, University of Cambridge, Trumpington Street,  Cambridge, CB2 1PZ, UK.
      }

\thanks{$^{4}$School of Mechanical Engineering, Northwestern Polytechnical University. 127 West Youyi Road, Beilin District, Xi'an Shaanxi, 710072, P.R. China.
        }
\thanks{$^{5}$Department of Computer Science, University of Aberdeen, AB24 3UE, Aberdeen, UK.
        {\tt\small e-mail: zhenhua.yu@abdn.ac.uk}}%
}
\begin{document}
\maketitle

\begin{abstract}

This paper proposes a wearable-controlled mobile manipulator system for intelligent smart home assistance, integrating MEMS capacitive microphones, IMU sensors, vibration motors, and pressure feedback to enhance human-robot interaction. The wearable device captures forearm muscle activity and converts it into real-time control signals for mobile manipulation. The wearable device achieves an offline classification accuracy of 88.33\%\ across six distinct movement-force classes for hand gestures by using a CNN-LSTM model, while real-world experiments involving five participants yield a practical accuracy of 83.33\%\ with an average system response time of 1.2 seconds. In Human-Robot synergy in navigation and grasping tasks, the robot achieved a 98\%\ task success rate with an average trajectory deviation of only 3.6 cm. Finally, the wearable-controlled mobile manipulator system achieved a 93.3\%\ gripping success rate, a transfer success of 95.6\%\, and a full-task success rate of 91.1\%\ during object grasping and transfer tests, in which a total of 9 object-texture combinations were evaluated. These three experiments' results validate the effectiveness of MEMS-based wearable sensing combined with multi-sensor fusion for reliable and intuitive control of assistive robots in smart home scenarios.
\end{abstract}


\section{INTRODUCTION}

\IEEEPARstart{T}{he} rapid development of smart home technology and the Internet of Things (IoT) \cite{9058658} has significantly increased the demand for intelligent and automated solutions in home environments. These solutions extend beyond entertainment, environmental control, and energy management, gradually expanding into home care and health monitoring. This trend is driven by the realities of global population aging, caregiver shortages, and rising healthcare costs, particularly in providing support for vulnerable groups such as individuals with disabilities, the elderly, and pregnant women \cite{IOT}. Intelligent health care technologies not only improve the quality of life for these populations but also significantly reduce the workload of caregivers. In the field of intelligent health care, the integration of mobile robots and wearable devices has gradually become a focal point, especially in the markets of the United States and Europe. Statistics from 2021 indicate that the annual cost of home care in the United States has exceeded 46.6 billion, and this figure continues to rise, further highlighting the increasing demand for care services \cite{health}.

Despite the significant potential of intelligent healthcare robots in assisting daily living and providing emergency monitoring\cite{10530645}, the current technological architecture of healthcare robots still lacks deep integration with other smart devices, particularly wearable sensors. Wearable devices have clear advantages in real-time health data monitoring and activity tracking \cite{9426433}, such as recording physiological indicators like heart rate, blood pressure \cite{heart}, and movement patterns\cite{activity}, and exchanging data with external devices \cite{application}. 
However, conventional healthcare robots fistly have not fully utilized this data, resulting in limitations in sensor system integration and cooperation, making it difficult to provide users with comprehensive, real-time health monitoring and feedback services. 
Furthermore, these robots often focus solely on their own sensing and control capabilities, lacking real-time interaction with wearable devices and full coverage of multi-sensor systems, which restricts their ability to achieve multifunctional cooperation. 
These limitations hinder the robots’ potential in real-world applications, especially in large-scale home environments where a high degree of real-time responsiveness and flexibility is required \cite{9510142}. 
Therefore, integrating technology to achieve a seamless connection between robots and wearable devices, thereby enhancing the real-time capabilities and flexibility of intelligent health care systems, has emerged as a significant research gap.

Current wearable devices mostly rely on EMG (electromyography), IMU (inertial measurement unit), and piezoelectric sensors to monitor muscle activity \cite{piezo}, but each of these sensors has its own limitations. EMG sensors are highly sensitive to external factors such as skin condition, electrode placement, and electromagnetic interference, resulting in significant signal quality degradation \cite{skin}. For instance, studies have shown that under non-ideal conditions such as electrode shift, muscle fatigue, physical friction, artifact noise, and sweating, EMG signal stability is significantly affected, leading to decreased signal accuracy \cite{EMG,EMG2}. Moreover, the complexity and high cost of EMG devices limit their widespread use in daily life. IMU sensors are widely used for motion monitoring;  however, they are prone to motion artifacts during fast or complex movements.  In particular, accuracy decreases in complex movements such as shoulder joint adduction-abduction and internal-external rotation, where the consistency between IMU measurements and Optical Motion Capture (OMC) systems is lower \cite{IMU}. Similarly, piezoelectric sensor signal quality often fluctuates with changes in contact pressure.
\begin{figure*}[ht!] 
\centering
\includegraphics[width=\textwidth]{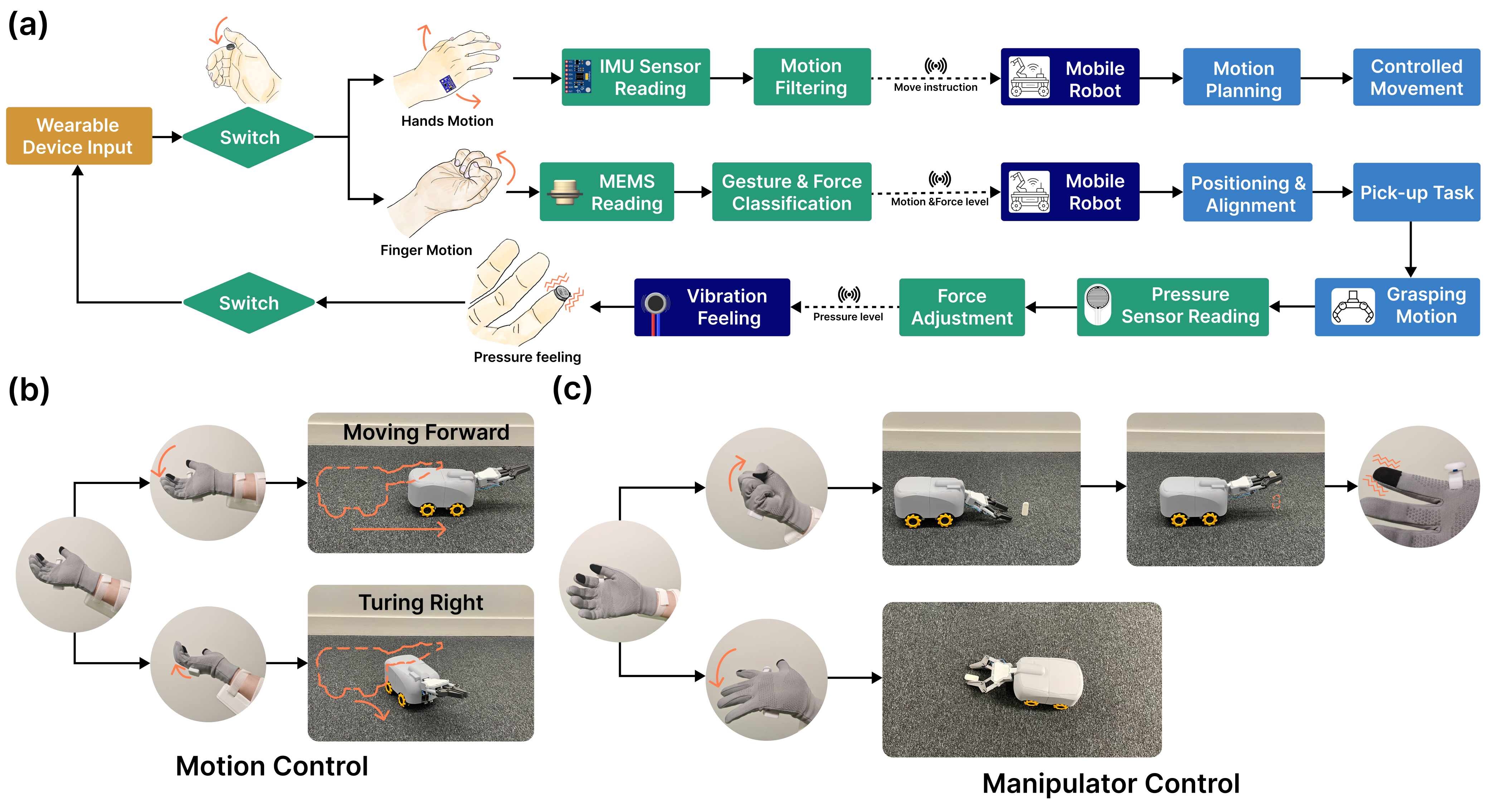}
\caption{
(a) Overview of the main operational flow between the wearable device and the mobile robot. The system uses a switch to select the task, and based on the acquired muscle signals and arm movements, it controls the mobile robot accordingly. This process ensures that commands are transmitted in real time, enabling effective human-robot interaction. (b) Illustration of the mobile robot's movement control. When the user's hand rotates forward or backward, the robot moves in the corresponding direction; similarly, lateral hand tilts induce left or right turns. This intuitive control mechanism allows the robot to navigate with precision in dynamic environments. (c) Depiction of the robot's grasping control. A finger-closing motion triggers the robot to grasp and pick up an object, whereas a finger-opening motion commands the robot to release the object or handle a square/box-like object. This adaptive grasping strategy supports the handling of objects with varying sizes, shapes, and textures.
}
\label{flow}
\end{figure*}
Visual recognition systems, which have been extensively used in the home healthcare monitoring, experience significant accuracy degradation in low-light environments or due to data privacy concerns \cite{ExLPose_2023_CVPR}. Studies indicate that in low-light conditions, the performance of many visual recognition algorithms can drop significantly \cite{light1}. Additionally, with the implementation of data privacy regulations such as GDPR and CCPA, visual systems face stricter constraints when collecting and processing user data, significantly reducing user acceptance \cite{Privacy}. 
Although microphone sensors can be used for sound recognition in home environments\cite{sound2}, noise interference remains a major limitation in practice. 
Research has shown that even in home settings, background noise can significantly reduce the accuracy of sound recognition, with error rates as high as 29.63$\%$ \cite{noise}.

In contrast, MEMS (Micro-Electro-Mechanical Systems) capacitive microphones perform exceptionally well in addressing these challenges. They exhibit high robustness in dealing with variations in conditions \cite{MEMS2} and electromagnetic interference \cite{ele}. Experimental data indicate that under the same environmental conditions, MEMS capacitive microphones demonstrate great signal stability, 97.73$\%$ identification accuracy is achieved on automatic classification through deep machine learning \cite{MEMS}. This makes MEMS capacitive microphones particularly advantageous and promising for applications in home environments and scenarios where EMG sensors under-perform.

Although extensive research and development efforts have focused on home health care robots \cite{old}, these systems primarily rely on visual sensors and remote sensing technologies. Such robots are often designed to assist elderly or disabled individuals with limited mobility by providing support for daily living activities and emergency monitoring \cite{care}. Furthermore, existing robotic systems offer limited interaction capabilities with users, particularly lacking sufficient support for real-time monitoring and feedback of muscle activity. Most robots rely on simple touch or voice interactions \cite{home} but fail to provide adequate real-time monitoring of users’ physiological states, particularly in capturing human activity data. This deficiency in interaction capabilities not only restricts the responsiveness and flexibility of robotic systems but also hinders their wider adoption and implementation in more complex home environments\cite{care2}.

To address these challenges, we propose a multi-sensor fusion-based remote-controlled mobile manipulator system. The system consists of two primary components: a wearable multi-sensor device worn by the user and a remote-controlled mobile manipulator robot. 
The wearable device captures the user’s muscle activity signals, subtle hand muscle movement and gesture, and hand force levels by combining the MEMS capacitive microphone with vibration motors, pressure sensors and IMU sensors. 
Through a multi-sensor fusion algorithm, these signals enable real-time control of the mobile manipulator, allowing the mobile manipulator to execute gripping tasks safely and accurately, minimizing errors due to imprecise control.
This system provides a more efficient solution for human-machine interaction in smart home settings, particularly in remote control and task execution.

In this study, we introduce, for the first time, a framework of integrating the multi-sensor fusion wearable with a mobile manipulator remote control for intelligent smart home assistance, expanding the ways in which humans interact with intelligent devices. This system is specifically designed for individuals with Parkinson's disease, gout, and other underlying health conditions, as well as for people with mobility impairments, such as pregnant women and individuals with disabilities, to assist them in completing everyday tasks at home. The main novelties of this research are:
\begin{enumerate}
    \item A multi-sensor fusion wearable integrates MEMS microphones, vibration motors, and IMU sensors to not only capture real-time muscle activity data and hand movement and gesture but also ensure precise control of the robotic arm in complex environments, mitigating the effects of electromagnetic noise and abrupt movements. 
    
    \item A real-time Human-Robot Interaction Framework which combines the wearable device with a CNN-LSTM machine learning model, the system enables real-time recognition of user gestures and force levels. This framework allows for precise and intuitive control of the mobile manipulator, facilitating efficient human-robot interaction in smart home applications.
\end{enumerate}

The structure of this paper is as follows: Section II proposes the design and test of the wearable device and mobile manipulator for experiments. Section IV explore the optimisation of the location of different sensors of
the wearable device and analyses the muscle activity signals of five forearm muscles for different muscle force by using a hybrid machine learning model. 
Three robot experiments test the effectiveness amd robutness of the proposed wearable-controlled mobile manipulators system in
Section V. Lastly, Section VI concludes the contribution of the paper.

\section{Hardware System Design}

The hardware system of the multi-sensor fusion-based mobile manipulator system consists of two primary subsystem: a multi-wearable device for hand control and a mobile manipulator. The integration of these two elements allows for real-time interaction between the user and the manipulator, enabling real-time control and feedback for intelligent task execution in smart home environments. 

\begin{figure}[htp!] 
\centering
\includegraphics[width=3.5in]{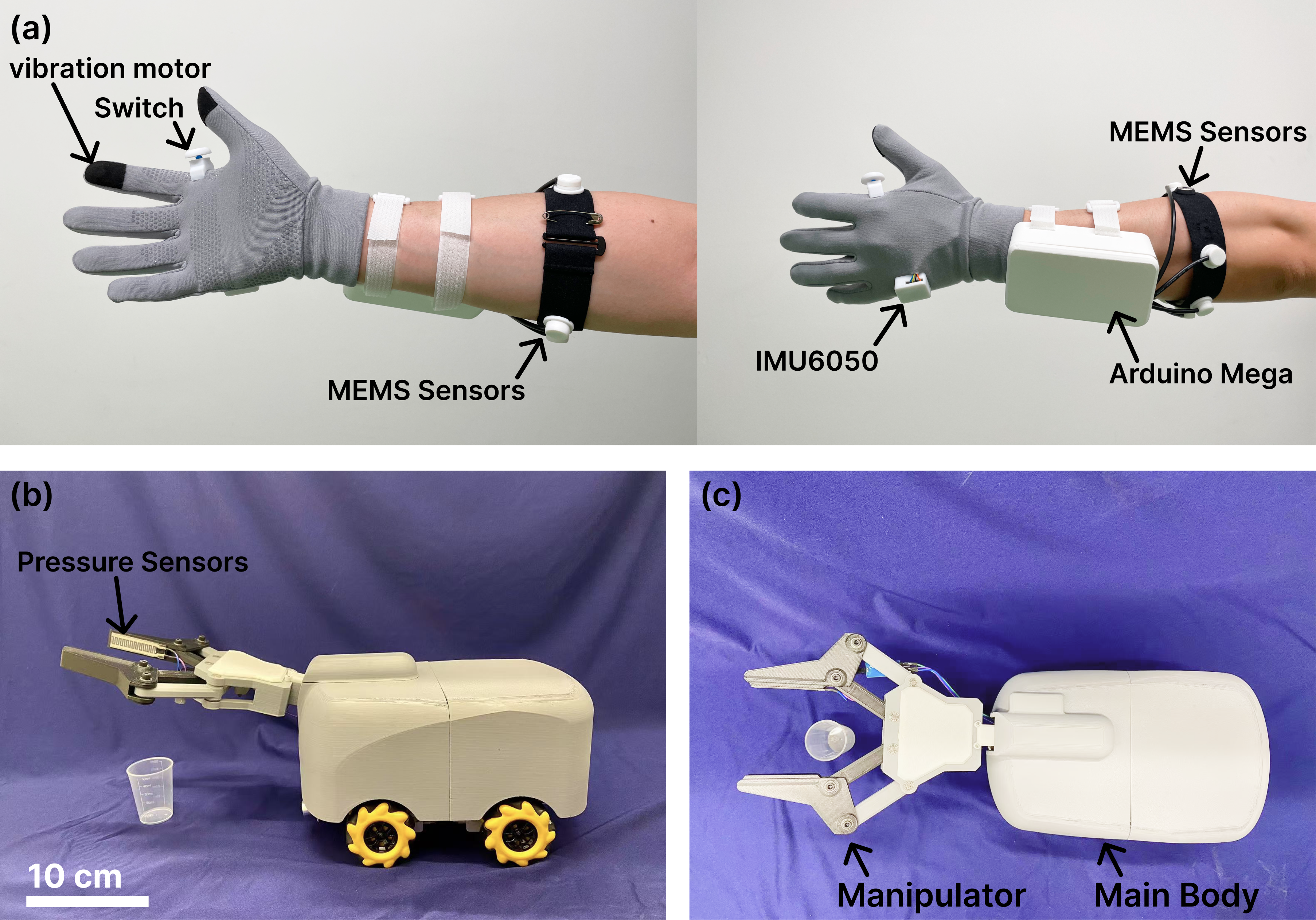}
\caption{
(a) The wearable device comprises three main components: a glove section featuring a vibration motor, IMU, and switch; a central module with an Arduino for signal processing; and a rear segment housing multiple MMG sensors for muscle activity detection.
(b) Side view of the mobile manipulator, highlighting the manipulator arm equipped with pressure sensors for force-based interaction.
(c) Top view of the mobile manipulator, illustrating its front-mounted manipulator section and a rear portion dedicated to motion control signal reception and processing.}
\label{device}
\end{figure}

\subsection{Wearable Device Design}

In this study, a wearable device was developed to detect human motion and provide mechanical feedback.   The hardware subsystem comprises three main components: a vibration motor and IMU sensor located on the wrist, a MEMS capacitive microphone sensor positioned at the forearm end, and an Arduino control module situated on the upper arm \ref{fig:sensor_placement}(b).   To enhance the accuracy of muscle movement detection, 5 MEMS capacitive microphone sensors are employed in a synchronized configuration.  According to relevant studies, the selected MEMS model demonstrates high sensitivity to low-frequency vibrations.


In the structural design of the MEMS sensor, particular attention was given to the reflection characteristics of sound waves within the cavity. A conical cavity structure (7 mm diameter, 5 mm height) was implemented at the base to focus sound waves, minimize reflections and attenuation, and enhance signal gain and sensitivity\cite{shape}. To further improve the acoustic performance, the rear cavity of the sensor was filled with Dragon Skin 30 silicone. This silicone material efficiently absorbs and dampens excess vibrations and noise, while also reducing interference from external sound waves\cite{sound}.

\begin{figure}[ht!] 
\centering
\includegraphics[width=3.3in]{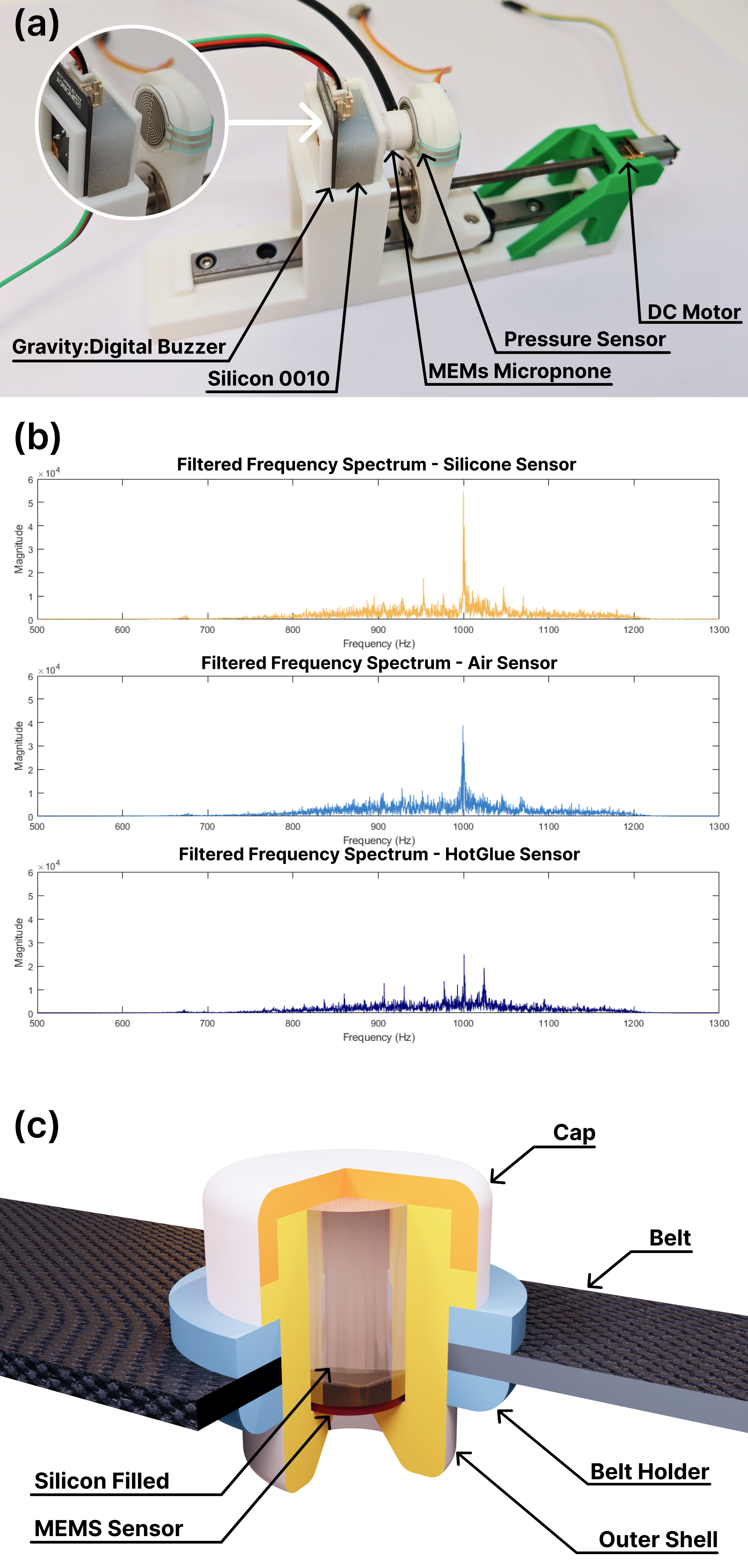}
\caption{
(a) Experimental setup designed for the MEMS microphone used to test different sensors under the same pressure and identical sound frequency conditions.
(b) Frequency response spectrum of three MEMS sensors using three kinds of material(Top: Silicone, Middle: Air, Buttom: Hot Glue) after band-pass filtering.
(c) Final internal sturcture of the MEMS sensor.
}
\label{sensor}
\end{figure}

To improve hand action recognition classification results, three sensor configurations with different filling materials in the sensor chamber cavity were tested: Dragon Skin 30 silicone-filled, hot melt adhesive-filled, and an empty cavity as reference.
Each sensor was subjected to controlled experimental conditions to simulate muscle vibrations. 
A custom-designed testing platform was used to standardize the testing conditions, ensuring consistent pressure and signal input across all sensors, as shown in Figure.\ref{sensor}(a). 
A Gravity: Digital Buzzer (1000Hz) was used to simulate muscle movement, with a 3mm silicon 0010 layer added to mimic human tissue properties. During the tests, consistent pressure was applied to each sensor using a motorized leadscrew and a pressure sensor, ensuring uniform pressure across all sensors at around 5N. The buzzer's input voltage was held constant at 3.3V, while the sampling frequency, controlled by Arduino, was set at 2600Hz to comply with the Nyquist theorem, thus preventing aliasing.

By analysing each sensor's response using Fast Fourier Transform (FFT), as shown in Figure.\ref{sensor}(b), the silicone-filled sensor provided the most robust performance, offering superior signal clarity and effective noise reduction when compared to the hot melt adhesive and empty cavity configurations.The silicone-filled sensor produced a well-defined peak at 1000Hz, corresponding exactly to the frequency generated by the buzzer, which also displayed minimal noise interference, confirming that Dragon Skin 30 silicone effectively dampened extraneous noise and retained the target frequency with high fidelity. 
In contrast, the hot melt adhesive-filled sensor presented a broader spectrum with less pronounced frequency peaks, indicating poorer damping performance. The FFT results exhibited additional noise surrounding the target 1000Hz signal, which suggests that the material was less effective in isolating the desired signal. 
Lastly, the empty cavity sensor showed the weakest performance. The frequency spectrum for this sensor displayed substantial noise across the entire range, with several unwanted peaks. This result highlighted significant interference, making it the least efficient configuration in terms of maintaining signal integrity.



Proper placement of the MEMS sensor is critical for signal accuracy and user comfort, minimizing motion artifacts and external interference during biosignal collection. In this study, the sensor was positioned on the forearm, approximately 4-5 cm below the elbow. This location ensures clear detection of muscle activity, particularly from the flexor muscles of the index and middle fingers. To guarantee sensor stability and user comfort, an adjustable elastic strap was used to secure the sensor in place, applying a controlled the force around 5N. Appropriate fixation pressure was crucial in maintaining effective signal transmission and reducing artifacts caused by sensor displacement  \cite{pressure}. Repeated experiments confirmed that this method of attachment effectively minimized sensor movement, ensuring consistent and accurate signal acquisition.

Both the IMU and vibration modules were mounted on the glove worn by the user, as shown in Figure \ref{device}(a). The IMU sensor was secured on the back of the hand, allowing the user to control the robot's forward, backward, and lateral movements through hand rotation. The vibration module was placed at the thumb tip for two reasons: (1) the thumb is minimally involved in muscle movements during hand rotation, reducing interference, and (2) its dense nerve network enhances sensitivity to vibration frequency changes. By placing the vibration module in this sensitive area, users can perceive vibration feedback more clearly, enabling more precise identification of frequency changes. This design optimizes vibration feedback perception while avoiding conflicts with other finger movements, ultimately enhancing both user experience and control precision.



\subsection{Mobile Manipulator Design}

The hardware of the mobile Manipulator primarily consists of two key components: the manipulator and the mobile chassis Figure.\ref{device}(b). 
The robotic arm is designed as a single-degree-of-freedom (DoF) system, equipped with a rotary joint and a two-finger manipulator as the end effector. This configuration allows for precise manipulation and ensures the safe handling of objects through stable gripping force. 

\begin{figure}[ht!] 
\centering
\includegraphics[width=3.5in]{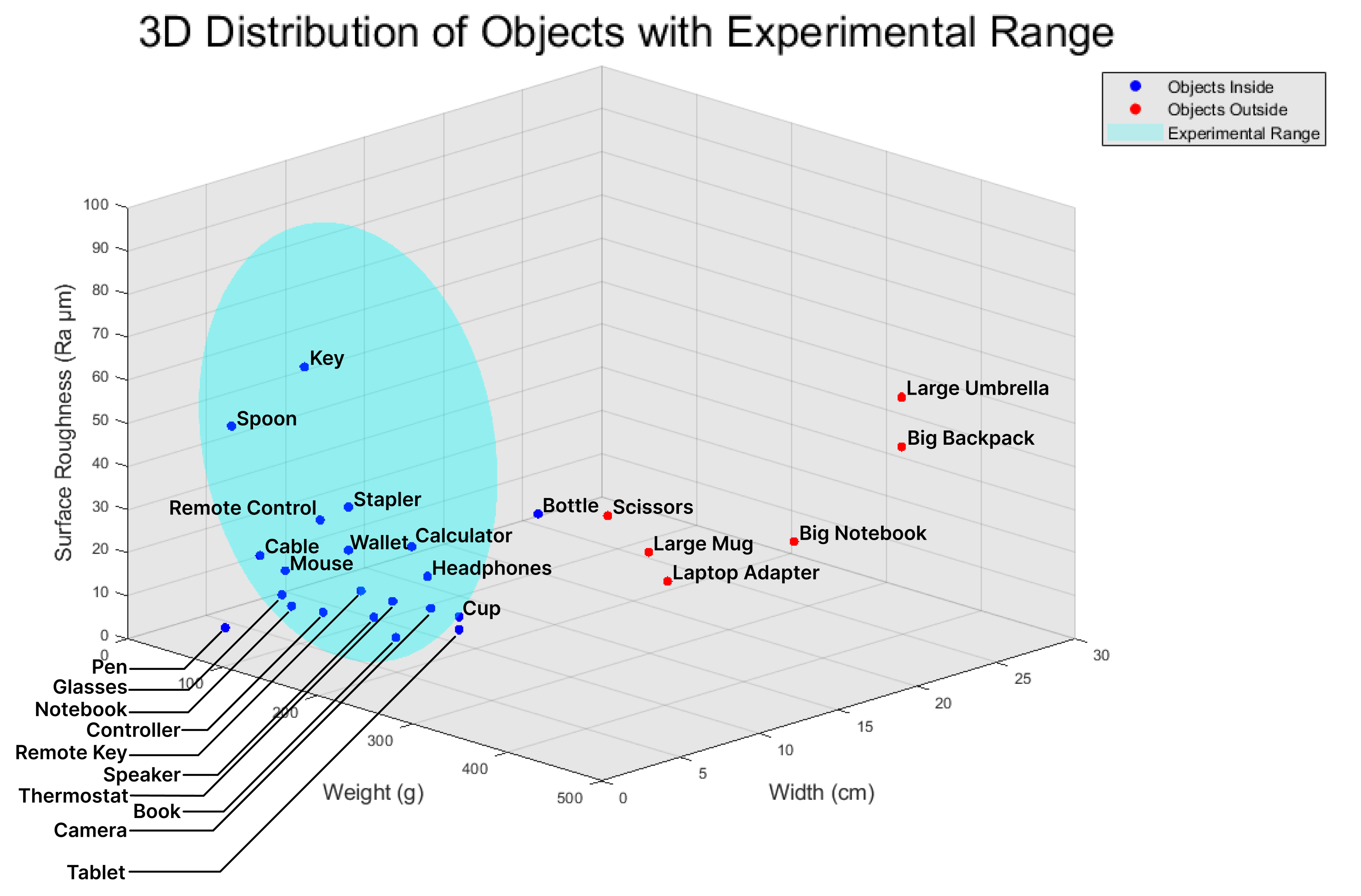}
\caption{3D scatter plot showing the distribution of common household objects based on weight (50-300g), width (cm), and surface roughness (Ra $\mu$m). Each point represents an individual item, providing a visual reference for the variability in these dimensions, aiding in the design and optimization of the robotic manipulator.}
\label{3D scatter plot}
\end{figure}

To accommodate the diversity of common household items, we conducted a study on typical everyday objects, identifying three key design dimensions: size, weight, and surface roughness. 
Figure.\ref{3D scatter plot}. presents a 3D scatter plot where each point represents a specific household item, such as a phone, remote control, cup, key, or glasses. The X-axis represents the weight of the object (Weight, g), the Y-axis represents its width (Width, cm), and the Z-axis indicates the surface roughness (Surface Roughness, Ra $\mu$m). The distribution of these points is based on actual measured data, and each point accurately reflects the properties of the objects in these three dimensions. 
The plot shows that most frequently used household items fall within the weight range of 50-300g, the width range of 3-7cm, and the surface roughness range of Ra 0-100$\mu$m. 
By observing the distribution of these points, clusters of objects with similar characteristics can be identified, such as keys that are lightweight and have rough surfaces, or cups that are heavier and smoother. 
These parameters guided the manipulator's design to efficiently handle most daily objects. Accordingly, the manipulator is set to have a length of 5cm, a maximum gripping diameter of 8cm, and a capacity to handle objects weighing up to 500g as shown in Figure. \ref{device}.

The mobile chassis uses a four-wheel drive system with Mecanum wheels, enabling omnidirectional movement, precise operations in confined spaces, and efficient navigation with a reduced turning radius, making it ideal for smart home applications.
An Arduino serves as the central control system, processing sensor data and executing control algorithms for precise manipulation and chassis movement. The system is powered by an 11.1V, 2200mAh lithium battery, ensuring stable operation and extended runtime.
For safety, the manipulator includes collision-avoidance sensors (ultrasonic and infrared) that trigger an emergency stop upon obstacle detection, along with a manual emergency stop button for immediate halting.
The force sensor in the gripper could provide real-time force monitoring due to the object surface roughness influencing gripping force requirements, particularly for delicate or irregular objects. It also provides force control feedback to the human through vibration motors in the wearable device.  

The manipulator features modular interfaces for future upgrades, such as adding camera modules for visual navigation or increasing the arm's degrees of freedom. The open-source Arduino platform enables advanced algorithms like optimized path planning and machine learning, aiming for greater autonomy and adaptability.

\subsection{Real-time Close Loop Control System}

When the user powers on both the wearable device and the robot, they can switch between two modes using a button near the thumb: movement mode (press and hold for 3s) and grasping mode (double press).

In movement mode, the IMU6050 sensor tracks hand tilts, converting real-time angle variations into robot movement commands sent via nRF24L01 wireless communication. The ultrasonic sensor on the robot detects obstacles and stops movement if needed, ensuring collision avoidance.

In grasping mode, the MEMS sensor captures forearm muscle signals to control the robotic arm. Data is wirelessly transmitted to a machine learning model, which identifies gesture type, force level (three levels), and surface roughness coefficient. The robot then executes the appropriate grasping action.

During grasping, the pressure sensor monitors applied force, adjusting it based on surface roughness estimation—smoother objects require stronger grip, while rougher objects need less force due to increased friction. The system quantifies grip force into six levels, mapped to eight vibration frequencies on the wearable device, allowing users to fine-tune their grip.

If the object is visibly identifiable, the user can apply a suitable force. Otherwise, they can start with a light grip and increase progressively based on vibration feedback, ensuring secure grasping while preventing damage.
For example, when handling a smooth, fragile object (e.g., a glass ball), the system detects slipping risk and prompts increased grip strength. Conversely, for a rough object (e.g., sandpaper), excessive force is unnecessary, and vibration feedback helps the user adjust accordingly. This adaptive haptic feedback system ensures precise, damage-free handling, even when surface properties are not visible.






\begin{figure*}[ht!] 
\centering
\includegraphics[width=\textwidth]{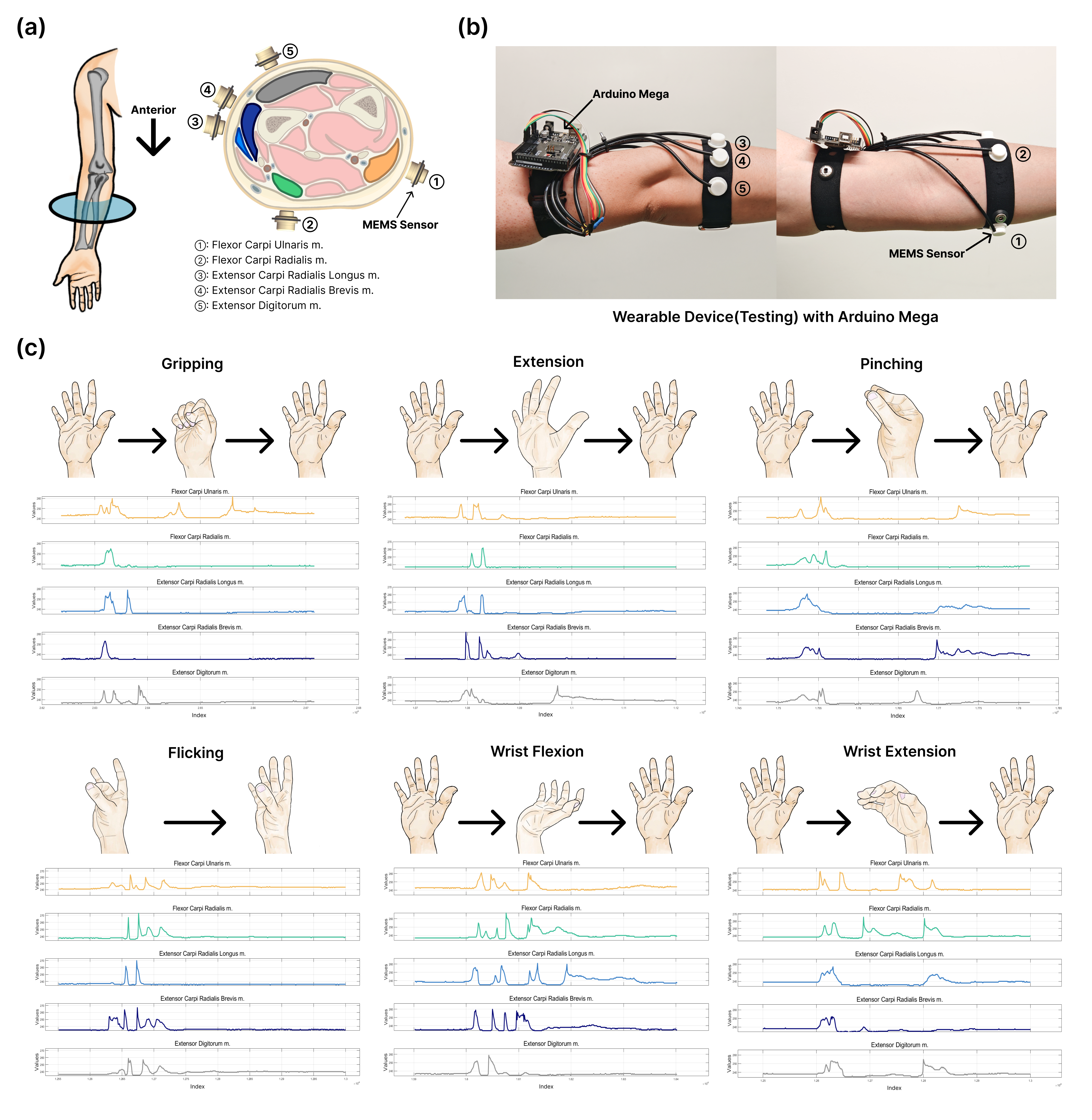}
\caption{(a) Cross-sectional diagram of the forearm showing the placement of MEMS sensors on five key muscles: Flexor Carpi Radialis, Flexor Carpi Ulnaris, Extensor Digitorum, Extensor Carpi Radialis Longus, and Extensor Carpi Radialis Brevis. The numbered labels indicate the sensor locations relative to each muscle, allowing for targeted signal detection during various hand and wrist movements.
(b) The wearable device used for testing consists of MEMS sensors and an Arduino Mega for data acquisition. The sensors are positioned on the forearm to capture muscle activity signals, while the Arduino Mega serves as the data collection and processing unit, ensuring real-time monitoring during different motion tasks.
(c) Basic characteristics of muscle signals during different movements, including gripping, finger extension, pinching, flicking, wrist flexion, and wrist extension. Each set of graphs corresponds to the signal responses from the five targeted muscles, illustrating how the amplitude and pattern of muscle activity vary depending on the type of movement performed.}
\label{fig:sensor_placement}
\end{figure*}

\section{Device Optimization and Muscle Activity Analysis}

This section optimizes the location of different sensors of the wearable device by analyzing the muscle activity signals of key forearm muscles. By analyzing the response patterns of muscles under different force levels, the system combines the CNN-LSTM machine learning model to classify grip force and wrist flexion force in real time, proving that our proposed design can accurately identify muscle activity characteristics to support applications in rehabilitation, prosthetic control, and human-computer interaction.

\subsection{Hand Gesture and Muscle Activity Analysis}
In the design of wearable devices, particularly those aimed at detecting and responding to hand muscle movements, determining the optimal placement of sensors is a critical step. As shown in Figure.~\ref{fig:sensor_placement}(a), this study focuses on identifying the most effective sensor locations on key forearm muscles: Flexor Carpi Radialis (FCR), Flexor Carpi Ulnaris(FCU), Extensor Digitorum(ED), Extensor Carpi Radialis Longus (ECRL), and Extensor Carpi Radialis Brevis(ECRB).
These muscles are primarily responsible for different hand and finger motions, making them essential targets for accurate sensor placement:
\begin{itemize}
    \item {Flexor Carpi Radialis (FCR)}: Controls wrist flexion and radial deviation (abduction toward the thumb).
    \item {Flexor Carpi Ulnaris (FCU)}: Responsible for wrist flexion and ulnar deviation (adduction toward the little finger).
    \item {Extensor Digitorum (ED)}: Facilitates finger extension, particularly of the second to fifth digits.
    \item {Extensor Carpi Radialis Longus (ECRL)}: Controls wrist extension and radial deviation, playing a key role in strong gripping motions.
    \item {Extensor Carpi Radialis Brevis (ECRB)}: Assists in wrist extension, especially during fine wrist movements and sustained gripping.
\end{itemize}
The primary objective of this study is to design a sensor layout that maximizes signal fidelity while minimizing user discomfort. By systematically evaluating different sensor positions, the goal is to enhance the wearable device's performance, ensuring it can accurately capture and process muscle signals in real-time. This is particularly important for applications such as human-machine interaction, rehabilitation, and prosthetic control. The ultimate aim is to establish a sensor arrangement that balances user comfort and signal clarity, meeting both practical usage needs and the precision required for detecting muscle activity.

As shown in Figure.~\ref{fig:sensor_placement}(b), repeated tests were conducted to identify which muscle movements produced the clearest and most stable data when detected by MEMS microphones. The pressure exerted by the sensors on the skin was carefully controlled and kept constant at (pressure value) throughout all tests to prevent variations in signal quality caused by inconsistent contact between the sensor and the skin surface. Additionally, irrelevant variables such as ambient temperature, participant posture, and sensor positioning angle were standardized to minimize any external influences on the sensor data, ensuring consistent and reliable results.

Daily hand movements, as shown in Figure.~\ref{fig:sensor_placement}(c), including gripping, finger extension, pinching, finger flicking, and wrist flexion and extension were selected as the focus of this experiment because they represent the most common and functional motions encountered in real-world scenarios. These activities activate the key forearm muscles in different ways, providing a comprehensive view of how each muscle contributes to everyday hand usage. 
For instance, the Flexor Carpi Radialis and Flexor Carpi Ulnaris are activated during gripping and wrist flexion, while the Extensor Digitorum plays a major role in finger extension. The Extensor Carpi Radialis Longus and Brevis are critical for wrist extension and radial deviation during tasks like gripping and lifting. 

The experiment focused on these natural movements to evaluate the sensors' real-world performance,  ensuring that the wearable device effectively captures meaningful data during routine activities. 
Emphasizing daily movements is crucial because the ultimate goal of wearable devices is to function seamlessly in everyday life, providing accurate feedback in real-time situations such as human-machine interaction, rehabilitation exercises, or prosthetic control.

Results showed distinct muscle activation patterns for different movements.  The Flexor Carpi Ulnaris and Extensor Digitorum muscles exhibited the most sensitivity to variations in finger gripping strength, with significant changes in the number and amplitude of peaks. Notably,  only the Flexor Carpi Ulnaris showed a distinct response when the hand was relaxed.
During finger extension, the muscle activity patterns were generally similar across all muscles, with two prominent peaks, particularly during rapid movement.
The Extensor Carpi Radialis Brevis differed significantly from gripping. 
In finger pinching, the Flexor Carpi Ulnaris and Extensor Carpi Radialis Brevis displayed clear responses,  but variations in finger position introduced noise into the data.

For the flicking motion, , as shown in Figure.~\ref{fig:sensor_placement}(c), all five muscles demonstrated similar signal characteristics, with Flexor Carpi Radialis, Extensor Carpi Radialis Longus, and Extensor Carpi Radialis Brevis showing significant changes at the same time points. However, repeated flicking led to noticeable thumb discomfort and finger fatigue, making the motion challenging to sustain. 
In wrist flexion, the Extensor Carpi Radialis Longus and Extensor Carpi Radialis Brevis generated the most distinct signals, with prominent peaks and minimal noise. 
In wrist extension, despite being less strenuous, yielded less defined signal characteristics from the Flexor Carpi Ulnaris and Flexor Carpi Radialis, and was more prone to interference from hand vibration. 
These results highlight the distinct muscle activation patterns and signal characteristics associated with different movements, underscoring the importance of optimizing sensor placement for specific hand motions.

Gripping tests revealed high responsiveness in Flexor Carpi Ulnaris and Extensor Digitorum to variations in grip strength, with significant differences in the number and amplitude of signal peaks, as shown in Figure.~\ref{fig:sensor_placement}(c),. 
The Flexor Carpi Ulnaris continued to show a distinct signal even when the hand was relaxed, while other muscles did not exhibit notable changes. This makes gripping an essential movement for sensor placement, as it is frequently used in daily tasks requiring different levels of force, and the corresponding muscles generate consistent and reliable signals.

Finger extension actived the Extensor Carpi Radialis Brevis and Extensor Digitorum muscles, which produced clear, distinct signals with pronounced peaks and minimal noise. The signal strength varied proportionally with the intensity of the movement, indicating their crucial role in wrist stabilization and force modulation. Although finger extension is less common in everyday activities compared to gripping, it plays a vital role in tasks requiring precise wrist control.

\subsection{Force-Dependent Muscle Activation Analysis }

\begin{figure}[ht!] 
\centering
\includegraphics[width=3.2in]{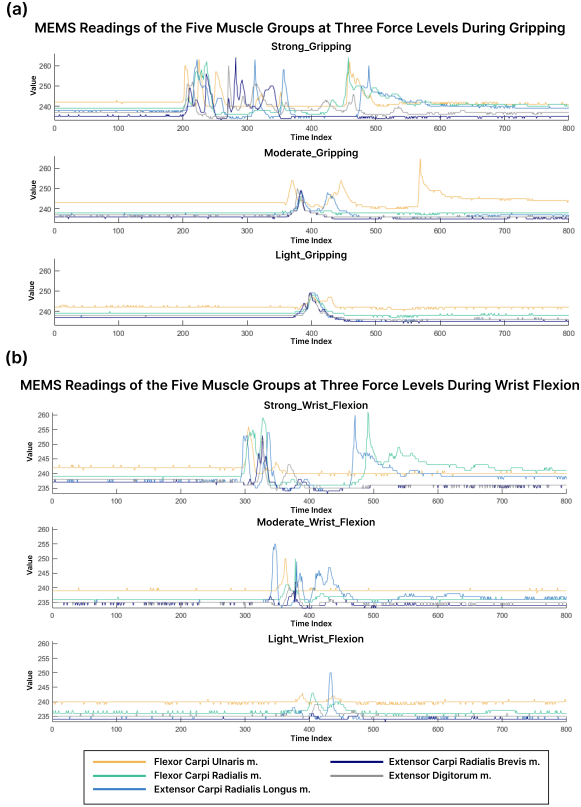}
\caption{MEMS muscle readings for six different force levels across two movements. From top to bottom, the graphs represent:Flexor Carpi Ulnaris; Flexor Carpi Radialis; Extensor Carpi Radialis Longus; Extensor Carpi Radialis Brevis; Extensor Digitorum. (a) MEMS readings at different force levels during gripping. (b) MEMS readings at different force levels during wrist flexion.}
\label{reservoir computing}
\end{figure}

Handling objects in daily life requires precise force control to prevent damage to fragile items while ensuring a secure grip on sturdier objects. 
While the previous subsection has demonstrated general muscle activation patterns during hand gesture movements, the specific variations in muscle response under different force intensities remain a challenge for wearable devices designed for force feedback applications. To address this gap, this subsection investigates the relationship between force intensity and forearm muscle activation during gripping and wrist flexion.

Preliminary experiments identified gripping and wrist flexion as two key hand movements essential for object manipulation. To mimic real-world scenarios, each was examined under three force levels: light, moderate, and strong. Fragile items require minimal force, while sturdier objects permit stronger grips. Object classification confirmed the need for adjustable grip control in daily tasks.

To quantify muscle activation under varying forces, MEMS sensors were placed on five forearm muscles. Data revealed distinct activation patterns corresponding to force intensity.

During gripping, higher forces led to greater fluctuations. At the strong level, the Flexor Carpi Ulnaris, Extensor Carpi Radialis Longus, and Brevis showed 3–4 high-amplitude peaks (up to 270 units), while moderate force produced ~2 peaks (~255 units), and light force yielded a single low peak (~235 units).

In wrist flexion, the Flexor Carpi Radialis and Extensor Carpi Radialis Longus dominated. Strong force caused 1–2 sharp peaks (~270 units), moderate force showed two moderate peaks (~250–260 units), and light force produced only one lower peak (~250 units).

These results highlight force-dependent activation: the Flexor Carpi Ulnaris was most sensitive during gripping, while the Flexor Carpi Radialis and Extensor Carpi Radialis Longus were key in both motions, especially under high force.

These findings establish a structured framework for classifying grip and wrist flexion force levels based on distinct muscle activation patterns.
The ability to differentiate force levels from muscle signals enables the development of real-time force feedback systems for wearable applications. 
By integrating machine learning algorithms, wearable devices can accurately recognize movement types and corresponding force intensities, allowing for precise force modulation in real-world scenarios. 

This approach enhances both safety and precision, ensuring that applied force consistently aligns with the specific handling requirements of different objects.

\subsection{CNN-LSTM Model for Force and Movement Classification}

This machine learning framework enables the wearable device to perform precise, real-time predictions, supporting practical applications where accurate force and movement recognition are essential. To enable the wearable device to perform precise, real-time predictions, a hybrid CNN-LSTM neural network was proposed to support practical applications where accurate force and movement recognition are essential. The MEMS sensor data, collected across different force levels and muscle groups, exhibited distinct time-series patterns, making it well-suited for sequential analysis. 
To enhance signal quality and mitigate noise, Savitzky-Golay filtering, defined by
\begin{equation}
y_i' = \sum_{j=-k}^{k} c_j \cdot y_{i+j}
\end{equation}
was applied as a preprocessing step, effectively smoothing the data while preserving key features such as peak amplitudes and signal trends—critical for distinguishing force levels. Here, $y_i'$ is the filtered output at index $i$, $y_{i+j}$ are the original signal values in a window of size $2k+1$, and $c_j$ are the polynomial-fitting coefficients.


Given the sequential nature of the data collected across different force levels and muscle groups, a hybrid CNN-LSTM neural network was employed, leveraging 1D convolutional layers to extract local spatial features and LSTM layers to capture long-term temporal dependencies. The data was further normalized using a mean-standard deviation transformation, ensuring consistency across different samples. Structured and preprocessed data was then fed into the CNN-LSTM model, which was trained with weighted cross-entropy loss to handle class imbalance effectively. By optimizing the model through adaptive learning rate scheduling, it achieved high classification accuracy for force intensities and movement types.

The model architecture consists of one 1D convolutional layer with 128 filters and a kernel size of 5, followed by a ReLU activation and reshaping to feed into a 5-layer LSTM with 256 hidden units per layer. The LSTM output is then passed through a fully connected layer to predict six output classes. The total number of trainable parameters is approximately 1.6 million, depending on the input dimensions.

The final classification is computed using the softmax function:
\begin{equation}
\hat{y}_i = \frac{\exp(z_i)}{\sum_{j=1}^{C} \exp(z_j)},
\end{equation}
where $z_i$ is the output logit of class $i$, $C$ is the number of classes, and $\hat{y}_i$ is the predicted probability. The predicted class is obtained via $\arg\max(\hat{y}_i)$.

This architecture allows for efficient representation of local sensor fluctuations and global temporal patterns, making it well-suited for classifying complex hand gestures and force levels.

\begin{figure}[ht!] 
\centering
\includegraphics[width=3.6in]{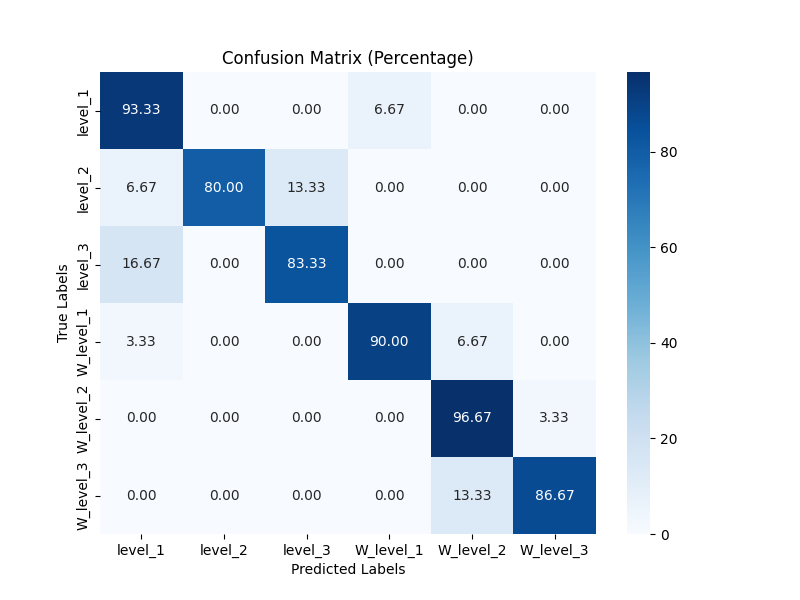}
\caption{
Confusion matrix representing the classification performance of the CNN-LSTM model for different force levels and movement types.
}
\label{cm}
\end{figure}

The confusion matrix as shown in Figure~\ref{cm}, illustrates the classification results of the CNN-LSTM model, demonstrating a high overall accuracy of 88.33$\%$. A key observation is that most misclassifications remain within the same general action category, indicating strong action recognition accuracy. Griping movements (level$\_$1, level$\_$2, level$\_$3) and extension movements (W$\_$level$\_$1, W$\_$level$\_$2, W$\_$level$\_$3) were accurately distinguished, with minimal cross-category misclassification. Higher force levels, particularly level$\_$1 and W$\_$level$\_$1, exhibited the highest accuracy, suggesting that muscle activation patterns for stronger forces are well-defined and easily separable. In contrast, medium and low force levels, such as level$\_$2 and level$\_$3, occasionally showed misclassification between adjacent force levels, which is expected given the subtle differences in muscle activation. Importantly, inter-category confusion, where stretching actions were mistaken for extension and vice versa, was minimal, reinforcing the model’s ability to differentiate between distinct movement types.

The hybrid CNN-LSTM model successfully captures both spatial and temporal signal characteristics, making it well-suited for real-time human- robot interaction applications.The incorporation of weighted cross-entropy loss effectively mitigates class imbalance issues, ensuring that recognition accuracy remains high across all movement types. 
These findings highlight the robustness of the machine learning framework, demonstrating that the wearable device can reliably interpret user intentions based on MEMS sensor data. 
By leveraging this classification system, the robotic control mechanism can dynamically adjust its grasping force and movement response based on real-time user inputs, enhancing the overall efficiency and precision of the human-robot interaction system.



\section{Human-Robot Application Performance Analysis}

To evaluate the effectiveness of the proposed wearable-controlled mobile manipulators system, a series of experiments were conducted to assess its performance in human-robot interaction. The analysis focuses on three key aspects: 1).gesture recognition precision under real-world conditions, which measures the system’s ability to correctly identify user gestures and force levels based on MEMS sensor data and a machine learning model; 2).human-robot synergy in navigation and grasping Tasks, which evaluates the probability of successfully reaching from Point A (initial position) to Point B (target location) while grasping an object, considering trajectory accuracy and environmental interactions; and 3).wearable-controlled mobile manipulators object transfer, which quantifies the system’s ability to execute stable and adaptive grasps across objects of varying sizes, shapes, and surface textures. These experiments provide a comprehensive assessment of the system’s robustness, adaptability, and real-world usability, offering insights into its feasibility for precise robotic manipulation and highlighting potential areas for further optimization.

\begin{figure*}[ht!] 
\centering
\includegraphics[width=\textwidth]{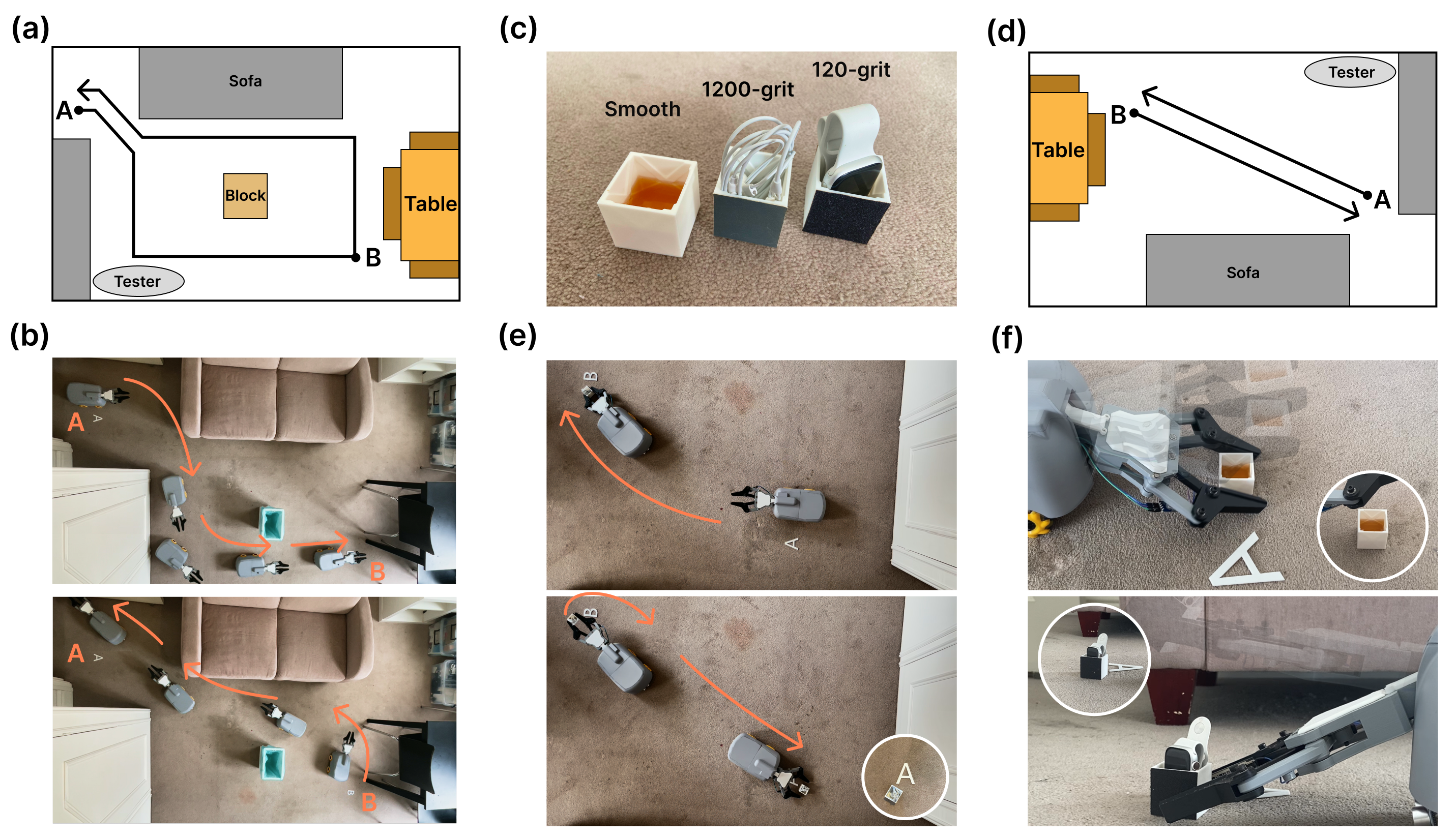}
\caption{
(a) Top-view schematic of the obstacle-rich transport path (Task 2), from Point A to Point B and back, navigating around furniture.
(b) Real-world implementation of Task 2, where the robot maneuvers around a sofa and central obstacle using wearable control.
(c) Test objects used in the transfer experiment: a water-filled cup with smooth surface, earphones with 1200-grit texture, and a watch with coarse 120-grit texture.
(d) Simplified transport path for Task 3, involving a direct point-to-point movement between A and B, used in the grasping and release evaluation.
(e) Execution of Task 3 in a real-world setting. The robot transports the object between designated locations based on surface texture and grip intensity.
(f) Detailed views of the releasing phase for different object types. The top image shows the stable placement of a water-filled object without spillage, demonstrating the system’s ability to adapt force and motion even for liquid-containing containers. The lower images highlight precise and secure release of dry textured objects.
}
\label{exp}
\end{figure*}

\subsection{Gesture Recognition Precision under Real-World Conditions}
To ensure precise control of the robotic system, the first experiment evaluated the accuracy of action classification based on MEMS sensor data. The ability to correctly identify different force levels and movement types is critical for reliable human-robot interaction. The classification model was trained and tested using a dataset that included multiple force levels across two movement categories: Gripping (level$\_$1, level$\_$2, level$\_$3) and Extension (level$\_$1, level$\_$2, level$\_$3). The CNN-LSTM model achieved an overall accuracy of 88.33$\%$, as shown in the confusion matrix (Figure~\ref{cm}). Although this result demonstrates strong performance in controlled settings, real-world applications may introduce challenges such as sensor noise, variability in user movement, and system latency that could affect the recognition accuracy.

To further validate the system’s reliability in practical use, a series of real-time experiments was conducted. Five test subjects (male and female, aged between 20 and 25) participated in these experiments. Each subject performed predefined movements while wearing the MEMS-equipped device, and the predicted classification results were compared against ground-truth labels obtained from motion tracking and manual annotations. For each force level, every action was repeated 50 times, and classification success was recorded based on whether the system correctly interpreted the user’s intended movement. Additionally, the response time from gesture execution to robotic action initiation was measured to evaluate real-time usability.


\begin{table}[htbp]
\centering
\resizebox{0.98\columnwidth}{!}{%
    \includegraphics[width=0.9\linewidth] {./figure/F23.pdf} 
}
\vspace{2mm}
\caption{Classification Accuracy and F1-Scores for Each Force Level in Gripping and Extension Movements.}
\label{tab:classification_accuracy}
\end{table}

The experimental results indicate that the MEMS sensor-based motion recognition system achieved an overall accuracy of 83.33$\%$, as shown in Table \ref{tab:classification_accuracy} in real-user testing, which is slightly lower than the 88.33$\%$ accuracy obtained by the CNN-LSTM model in offline training. This discrepancy may be attributed to factors such as sensor noise, individual variations, and system latency. The average recognition response time was 1.2 seconds, with some cases reaching 1.3 seconds. Notably, the classification accuracy for Gripping-Level 2 and Extension-Level 3 was relatively lower, likely due to high similarity in motion characteristics, sensor sensitivity limitations, and differences in individual movement patterns.  Overall, the system demonstrates strong potential for human-robot interaction applications but requires further optimization to ensure greater stability and practicality.

\subsection{Human-Robot Synergy in Navigation and Grasping Tasks}
The second experiment evaluates the success rate of reaching and grasping tasks, focusing on trajectory precision, movement smoothness, and ease of control to assess the overall human-robot interaction efficiency.  As the system is designed for real-time wearable control, the user's ability to intuitively guide the robot plays a crucial role in determining task success.  The effectiveness of the control scheme is reflected in how easily users can maneuver the robot, how accurately it follows intended paths, and how consistently it reaches and grasps target objects.  Deviations in trajectory, unintended movements, or excessive corrections can increase task difficulty, leading to inefficiencies in execution.

The experiment was conducted in a controlled home environment designed to simulate a realistic indoor setting. The space was furnished with tables, chairs, and cabinets that naturally introduced obstacles along the robot's path (see Figure~\ref{exp} (a)). The predefined trajectory spanned a total of 8 meters, incorporating three 90-degree turns and two 45-degree turns, which required users to actively adjust their movement commands. The primary objective was to evaluate the accuracy and reliability of the wearable device's control over the robot's movement as it navigated from Point A (the starting position) to Point B (the target object location) and go back to Point A.

\begin{table}[htbp]
\centering
\resizebox{0.98\columnwidth}{!}{%
    \includegraphics[width=0.9\linewidth]{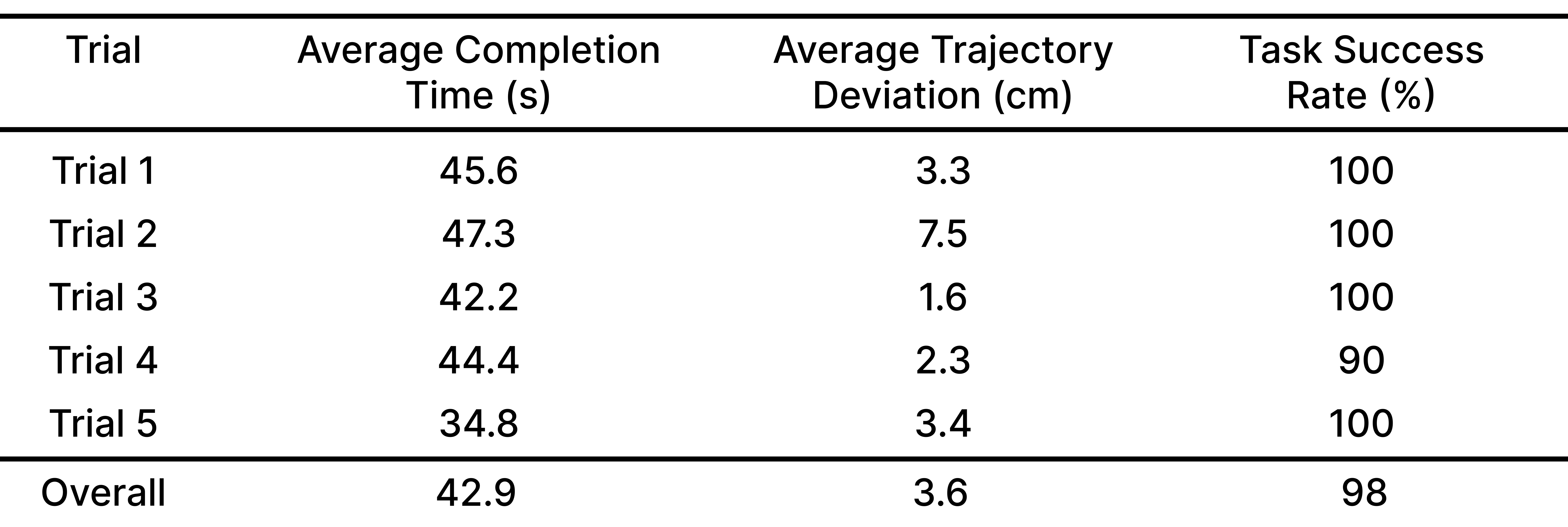} 
}
\vspace{1mm}
\caption{Performance Metrics of the Wearable-Controlled Mobile Manipulator}
\label{tab:performance_indicators}
\end{table}

In the experiment, five test subjects were recruited to evaluate the wearable-controlled mobile robot. Each subject was required to use the wearable device to navigate the robot from Point A to Point B and go back to A, with a trial considered successful if no collisions occurred as shown in Figure \ref{exp} (b). Every subject completed 10 trials, and key performance indicators—Average Completion Time(s), Average Trajectory Deviation(cm), and Task Success Rate—were computed for each trial. Although each trial took a relatively long duration, the overall success rate was extremely high, with only one collision observed out of 100 trials as shown in Table \ref{tab:performance_indicators} . This single failure occurred when the robot hit the cabinet while turning.  On average, subjects completed the movement in about 30 seconds, and the final distance to the target was typically less than 4 cm, which is within the grasping range of the robotic gripper. Given additional time or the opportunity for repeated attempts, it is expected that fine-tuning during the final stage could further improve the success of target object grasping.

Overall, the system demonstrated extremely high reliability and accuracy, indicating that with additional time or repeated fine adjustments, the wearable-controlled robot could consistently perform precise object manipulation tasks in realistic scenarios.


\subsection{ Wearable-Controlled Mobile Manipulators Object Transfer}

This experiment aimed to evaluate the grasping and transportation capabilities of a MEMS-based wearable system, focusing on the user's ability to adjust grip strength based on real-time haptic feedback.   Unlike previous trials that primarily emphasized trajectory tracking or robotic control accuracy, this study investigated how human users interpret surface roughness through vibration cues and adapt their grip force accordingly during real-world object manipulation.

The experiment consisted of three continuous phases: grasping, transporting, and releasing.   Four everyday objects were used—namely a watch, earphones, a small water cup—with each presented in one of three surface conditions: smooth, fine (1200-grit), and coarse (120-grit) as shown in figure \ref{exp} (c). These combinations were randomly placed inside identical boxes to introduce variability in handling.   Two participants each completed two trials for every object-texture pairing, resulting in a diverse and comprehensive dataset.

In the grasping phase, users first observed the object visually and selected an appropriate grip level (Level 1, 2, or 3) based on its estimated weight and surface properties as shown in Figure \ref{exp} (d). A pressure sensor embedded in the manipulator monitored the actual applied force, while a MEMS sensor worn on the user’s forearm captured muscle signals, which were mapped to grip intensity through a calibrated model. Simultaneously, real-time vibration feedback indicating surface roughness was delivered to the user, allowing them to adjust their grip strength intuitively—rougher textures typically required less force due to increased friction.

The transport phase required users to carry the object from Point A to Point B as shown in Figure \ref{exp} (e). During this process, the object had to remain intact and stable, and the manipulator was not allowed to collide with any obstacles or make scraping contact with the environment. As shown in Table \ref{tab:experiment_conclusion}, recorded transport times were consistently under 40 seconds, with noticeable reductions in time as participants became more familiar with the task.   Most of the time was spent during turning maneuvers, while grasping and releasing took relatively little time.   Users demonstrated high success rates in moving even heavier objects like watches and tablets.   However, some failures were observed—particularly in the water cup with a smooth surface case—where one user misjudged the cup’s weight and applied insufficient grip strength, resulting in a failed lift.

\begin{table}[htbp]
\centering
\resizebox{0.98\columnwidth}{!}{%
    \includegraphics[width=0.9\linewidth]{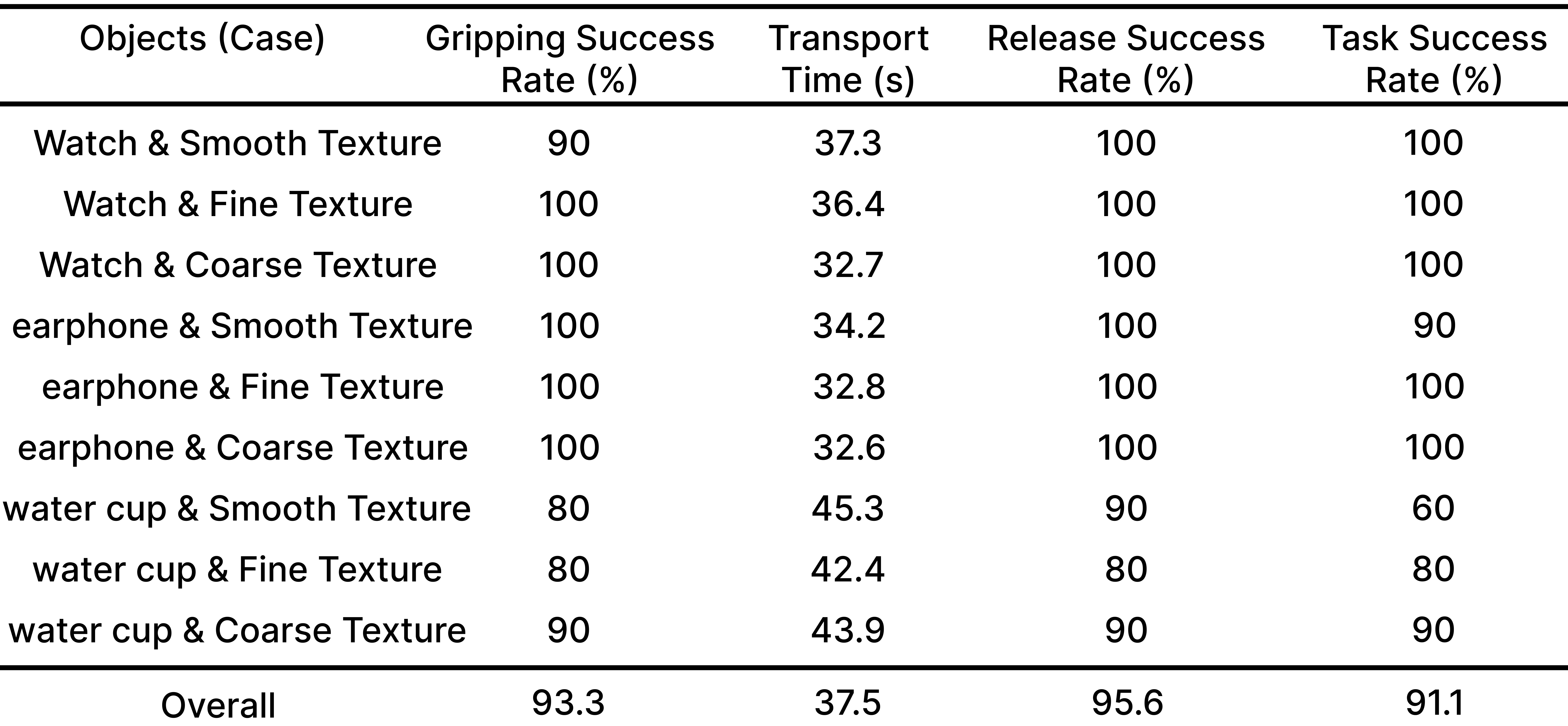} 
}
\vspace{1mm}
\caption{Performance Metrics of the Wearable-Controlled Mobile Manipulator}
\label{tab:experiment_conclusion}
\end{table}
During release at Point B as shown in Figure \ref{exp} (f), users were allowed to choose among three strategies: light, standard, or gradual release. The system monitored whether the object was placed stably and whether any damage or spilling occurred. For the water cup, all participants chose a gentle release strategy to prevent spilling.   While no water spilled during the actual release phase, one incident of water overflow occurred during transport due to excessive movement during turning, suggesting that user handling habits still influence task outcomes despite proper grip control.

Overall, the MEMS-controlled system enabled users to successfully complete object manipulation tasks with high precision and adaptability.   Participants were generally able to interpret vibration feedback to infer surface textures and adjust grip forces appropriately.   Failures primarily stemmed from initial misjudgment or overly aggressive movement, not from system limitations.   These findings validate the system’s effectiveness in real-world human–machine interaction scenarios that require dynamic force modulation and real-time sensory feedback integration.



\section{Discussion \& Conclusion}

In this work, we proposed and validated a multi-sensor fusion-based wearable control system for mobile manipulators, specifically tailored for intelligent assistance in smart home environments.  By integrating MEMS capacitive microphones, IMU, and pressure sensors, the wearable device enables robust sensing of forearm muscle activity, which is used to control a mobile manipulator in real-time.  
Offline machine learning using a CNN-LSTM architecture achieved 88.33\%\ classification accuracy across six motion-force levels, while real-time user tests maintained an accuracy of 83.33\%, confirming system feasibility in practical use.

Our hardware experiments further demonstrated the robot’s ability to execute smooth, collision-free movement with an average trajectory deviation of 3.6 cm and a task completion rate of 98\%\ in a realistic indoor path. During object manipulation experiments, 12 distinct object-texture combinations were tested. Results show a gripping success rate of 93.3\%, a release success rate of 95.6\%, and a full-task success rate of 91.1\%. 
Notably, the system enabled users to adjust their grip according to surface roughness using vibration feedback, and most grasping errors occurred due to initial underestimation of object weight rather than system failure. These findings highlight the practical potential of MEMS-based wearable systems in augmenting human-robot collaboration.  Future work will explore expanding the system’s motion vocabulary, improving grip-force precision through adaptive learning, and integrating visual perception to further enhance autonomy and environmental adaptability.

\addtolength{\textheight}{-12cm}   
 
\bibliographystyle{IEEEtran}
\bibliography{bibs}
\end{document}